\documentclass[10pt,twocolumn,letterpaper]{article}

\usepackage{wacv}
\usepackage{times}
\usepackage{epsfig}
\usepackage{graphicx}
\usepackage{amsmath}
\usepackage{amssymb}
\usepackage{booktabs}
\usepackage{algorithm}
\usepackage{algorithmic}

\wacvfinalcopy 

\def\wacvPaperID{****} 

\ifwacvfinal\fi
\begin{document}

\title{Cooperative Initialization based Deep Neural Network Training}

\author{Pravendra Singh \hspace{2cm} Munender Varshney \hspace{2cm} Vinay P. Namboodiri\\
Department of Computer Science and Engineering, IIT Kanpur, India\\
{\tt\small \{psingh, munender, vinaypn\}@iitk.ac.in}
}

\maketitle
\ifwacvfinal\thispagestyle{empty}\fi

\begin{abstract}
Researchers have proposed various activation functions. These activation functions help the deep network to learn non-linear behavior with a significant effect on training dynamics and task performance. The performance of these activations also depends on the initial state of the weight parameters, i.e., different initial state leads to a difference in the performance of a network. In this paper, we have proposed a cooperative initialization for training the deep network using ReLU activation function to improve the network performance. Our approach uses multiple activation functions in the initial few epochs for the update of all sets of weight parameters while training the network. These activation functions cooperate to overcome their drawbacks in the update of weight parameters, which in effect learn better ``feature representation" and boost the network performance later. Cooperative initialization based training also helps in reducing the overfitting problem and does not increase the number of parameters, inference (test) time in the final model while improving the performance. Experiments show that our approach outperforms various baselines and, at the same time, performs well over various tasks such as classification and detection. The Top-1 classification accuracy of the model trained using our approach improves by 2.8\% for VGG-16 and 2.1\% for ResNet-56 on CIFAR-100 dataset.
\end{abstract}

\section{Introduction} 
Deep neural networks (DNNs) are state-of-the-art models, responsible for transforming research in the area of vision, language and speech \cite{lecun2015deep,graves2013speech,collobert2008unified}.
Various works \cite{singh2019hetconvijcv,singh2019hetconv,singh2019stability,singh2018leveraging,singh2019multi,singh2019accuracy,singh2019falf,mazumder2019cpwc,singh2019play} have been proposed for efficient deep learning. These deep network at core performs a linear transformation followed by a non-linear operation using an activation function. The activation function is the one, which is responsible for nonlinear behaviour and the learning capabilities of the network. These activations are
non-linear continuous functions which may also possess non-differentiability \cite{nair2010rectified,maas2013rectifier,he2015delving}. Researchers have proposed many activation functions which can be classified into saturated \cite{bishop1995neural,mcculloch1943logical,hinton2009replicated} and non-saturated activation functions \cite{bishop1995neural,maas2013rectifier,xu2015empirical,nair2010rectified}. 

The saturated activation belongs to a category, in which the learning process gets slow down due to the very small gradient near-saturated output. These activation functions are experimentally proved to be less effective for training a deep network. The key reason for the failure is the (vanishing/exploding) gradient problems, which mostly occurs due to saturated output in an activation function. This problem is  efficiently tackled by using non-saturated activation function, like ReLU \cite{nair2010rectified,maas2013rectifier}. In particular, the derivative of ReLU is one for the positive inputs; hence, the gradient cannot vanish. In contrast, all the negative values are mapped to zero, which restricts the flow of information in DNNs for these negative values. ReLU gets saturated exactly at zero, which makes ReLU fragile at the time of training, and the neuron can die forever. For example, the flow of large gradients through ReLU may update weight parameters in a way that may deactivate neurons for all data points. This problem is known as dying ReLU, which implies that the gradients flow through the neuron will forever be zero from that point. Due to this, the gradient-based optimization algorithm will not be able to update the weights of that neuron unit. Also, training the network on a high learning rate may shoot the number of ``dead" neurons in the network as much as 40\% of the network \cite{maas2013rectifier} (i.e., neurons that never activate across the entire training dataset). So there is a need to set the learning rate properly to reduce the issues.

To resolve these potential problems originated by the hard zero mappings in the ReLU units, various generalizations of ReLU such as Leaky ReLU \cite{maas2013rectifier}, and PReLU \cite{he2015delving} have been proposed. Both Leaky ReLU and PReLU are same as ReLU except for the case of negative inputs in which a small constant slope for Leaky ReLU and a learnable slope for PReLU are used. Similarly, Exponential Linear Units (ELU) \cite{clevert2015fast} is also proposed, which resolves the bias shift \cite{ioffe2015batch} from the succeeding layers. ELU \cite{clevert2015fast} gives an exponential value corresponding to negative inputs, which force the mean output of the activation function to reach towards zero. Although ELU is not backed by concrete theory, ELU shows competitive results. Besides all these, Softplus \cite{zheng2015improving} is approximately similar to ReLU, except at zero, where softplus is differentiable and smooth. Softplus is also differentiable everywhere and saturates less, which gives an edge over ReLU. In practice, there is no non-linear activation function that outperforms all the time over all models, datasets, and problems.

In this paper, we propose an approach in which multiple non-linear activation functions are exploited using a cooperative strategy to overcome their drawbacks. We have used multiple activation functions in the initial epochs of training the deep network. The aggregation of gradient from all the activation function gives a \textit{regularization effect} for the gradient flow corresponding to the whole range of inputs (negative values also). This results in regularizing the update of weight parameters, which is a very crucial step in the initial stage. In the next stage, we train the network with only one activation function (standard network) corresponding to each layer of the network. The proposed approach has experimented extensively on different architectures such as ResNet, and VGG-16 models over CIFAR-10, CIFAR-100, and ImageNet datasets. We also experimented with object detection task using SSD-300 on the PASCAL VOC dataset.

Major contributions of this paper are as follows:
\begin{itemize}
    \item  We have shown experimentally that using multiple activation functions in the initial few epochs of the training process benefits the update of the full set of weight parameters, which results in substantial performance improvement later.
    \item  We have shown empirically that a mixture of non-linear activation functions results in significant improvement in the performance as compared to the individual non-linear activation function.
    \item We have shown that Cooperative Initialization based training also help in reducing overfitting problem.
    \item Our proposed approach does not increase the number of parameters and inference (test) time in the final model while improving the performance.
\end{itemize}

\section{Previous Work}

The first activation function is a step function originally used in the perceptron model \cite{mcculloch1943logical}. Researchers in the same direction have also proposed many other saturated activation functions such as sigmoid, softmax, and tanh \cite{bishop1995neural}. The ReLU activation further replaces these activation functions, owing to the outstanding performance on deep neural networks \cite{nair2010rectified,szegedy2015going,krizhevsky2012imagenet}. ReLU had escalated the convergence and resolved the vanish gradient problem, normally occurred in saturated activation functions. These activation functions have accelerated the efforts of the research community in solving various vision problems. Several attempts have been made to develop a more efficient network by developing better activation functions, which can resolve the problems arising in the above activation functions. Some variants of ReLU have been proposed such as RReLU, PReLU, leaky ReLU, and others \cite{xu2015empirical,he2015delving,maas2013rectifier}. 

To resolve the issues of mapping, all negative input to zero (dying ReLU) in the ReLU activation function, Leaky ReLU \cite{maas2013rectifier} has been proposed. This mapping causes an information loss (dead neuron), which is resolved by defining a linear function corresponding to negative input, having a small predefined constant slope, to leak some information \cite{maas2013rectifier}.  However, Leaky ReLU does not give any notable results on performance experimentally. Further, a parametric rectified linear unit (PReLU) has been proposed \cite{he2015delving} which uses a learnable slope  parameter instead of a constant slope as in Leaky ReLU \cite{maas2013rectifier} for negative inputs. PReLU gives better performance than ReLU in many cases. On the second thought, the slope parameter can be randomly sampled from a uniform distribution as used in Randomized Leaky Rectified Linear Unit (RReLU) \cite{xu2015empirical} which reduces the risk of overfitting in the training phase. ELU \cite{clevert2015fast} is also same as ReLU for positive input while it behaves similarly to saturated exponential function for negative inputs. Further, Parametric ELU (PELU) is a scaled version of ELU having a learnable scaling parameter \cite{trottier2017parametric}. Most of the above discussed ReLU variants are based on the experiments over the negative inputs of ReLU.  

There are few other works in which new activation functions are proposed, such as Maxout (maximum over K affine functions), Softplus, and Adaptive Piecewise Linear (APL) \cite{goodfellow2013maxout,zheng2015improving,agostinelli2014learning}. The APL consist of many non-differentiable points that scale linearly with the number of hinge functions. This will increase the model complexity and affect the parameter updates during back-propagation \cite{lecun2015deep}. In the same manner, Maxout take maximum over multiple feature maps. Softplus is the smooth approximation of ReLU, which is differentiable everywhere. 

 In this work, we have focused on leveraging benefits from the multiple non-linear activation functions simultaneously. To the best of our knowledge, this is the first work that considers a mixture of non-linear activation functions in the initial few epochs. We have also presented an ablation study and feature visualization to support the proposed approach.   

\begin{figure*}[!t]
\includegraphics[scale=0.42]{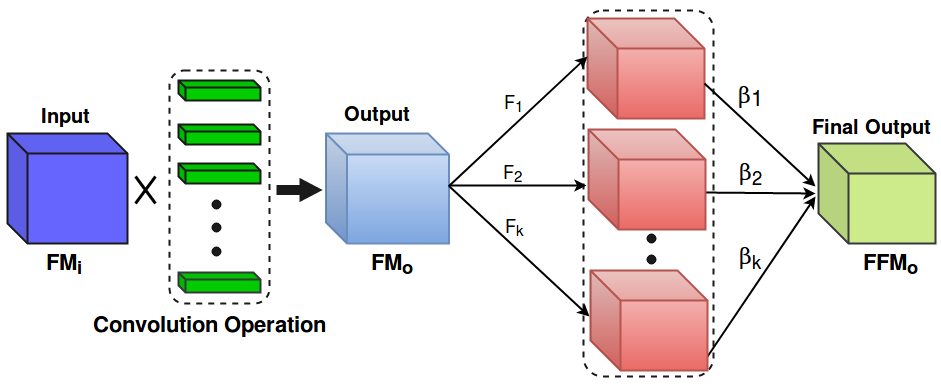}
\centering
\caption{Cooperative design for a particular layer in deep network during Phase-1 training time (Best viewed in color).}
\vspace{-15pt}
\label{netdesign} 
\end{figure*}

\section{Cooperative Network Design (Phase-1)}

Nowadays, deep networks consist of huge depth due to the presence of multiple convolutional layers. These convolutional layers are followed by an activation function, which operates on the feature maps (output) of these layers. The training process will update weight parameters using back-propagation. This update solely depends on the behavior of the activation function, i.e., ReLU never updates the weight parameter corresponding to negative inputs. Restricting these weights from an update in the initial phase of training may hurt the performance of the deep network later. 

To overcome these issues in the network, researchers have proposed various activation functions, which have their advantages and disadvantages. In this work, we have proposed a \textit{cooperative design} having multiple activation functions, helping each other in the initial update of parameters in such a way that all set of weight parameters get an opportunity to contribute to the performance. These activation functions cooperate to overcome their drawbacks and improve the update process for all sets of weight parameters in the initial epochs of training. 

In Figure \ref{netdesign}, we have shown a block diagram of a layer, where the convolution operation is same as of standard CNN, but we have used k activation functions instead of one activation function. The input feature maps for the given layer are represented as $FM_i$ in Figure \ref{netdesign}, which are then convolved using convolutional filters shown in green color. The output feature maps $FM_o$ generated by convolution operation are passed to each activation function, which operates over each element of the given feature maps. This results in the generation of k different sets of feature maps corresponding to each activation function, as shown in Figure \ref{netdesign}. The Final output Feature Maps ($FFM_o$) are the result of the weighted average of these k feature maps as given by the following equation:

\[  FFM_o = \beta_1 F_1(FM_o) + \beta_2 F_2(FM_o) + ... + \beta_k F_k(FM_o) ; \]
where $F_1, F_2, ...  F_k $ are the activation functions applied on the $FM_o$ feature map. The final output feature maps ($FFM_o$) are the weighted average of outputs from these activation function when applied to input feature maps (shown in red color). The corresponding weights for averaging are $\beta_1, \beta_2, ..., \beta_k$ where $1,2, ..., k$ represents the k activation functions. We assume that each activation contributes equally in the update process (improvement) of the weight parameters by assigning equal weight to each parameter ($ \beta_i = 1/k $).  

We have presented an ablation study to shown empirically that even if we train the model completely by using a mixture of activation functions, it will results in substantial performance improvement as compare to the individual activation function. 

\section{Standard Network Design (Phase-2)}
In Phase-2, the network use only a single activation function, i.e., ReLU activation for each layer in the model, instead of the mixture of activation functions. In the training process of phase-1 design, our focus is to shift the model in a stable state where all sets of model parameters are in a better state than a randomly initialized state. The phase-1 network design is trained for a few epochs such that the mixture of activation functions will overcome the drawbacks of each other by cooperation in the update process of weight parameters. All the k activation functions will get different gradients and averaging them give a \textit{regularization effect}, which updates all sets of parameters uniformly without undermining others. 

In Phase-2, We use only one activation function, so we opt to use only ReLU as our activation function. ReLU is a linear (identity) mapping for all positive values and zero values for all negative values. ReLU activation shows a sparse behavior as all negative inputs are mapped to zero. After getting into a better state from phase-1, we often desire to make feature maps sparse enough, which is easily possible with ReLU as it possesses many deactivated neurons, giving a regularized effect to the model. Sparsity results in concise models that often have better predictive power and fewer chances of overfitting/noise. In a sparse network, all neurons cannot be activated simultaneously in a model. Only those set of neurons get activated which are responsible for a particular aspect of the given task, e.g., there are a given set of neurons which gets activated for a face like structure in a human detecting task while the same set of neurons are inactive for other parts of the body. That's the reason standard ReLU seems to be less prone to overfitting vs. leaky ReLU with modern architectures. We have also presented an ablation study to empirically show that ReLU in Phase-2 will results in substantially improved performance as compare to other possible options. Hence ReLU is competing with all other non-linear activation functions in Phase-2.

\section{Training Method}

The proposed training framework is divided into two phases corresponding to the two network design phases. For Phase-1 (Cooperative) training, we use 20\% of epochs used in standard (Phase-2) training with a mixture of non-linear activation functions (having equal weight).
We have presented an ablation on the number of epochs used in Phase-1 training to validate the choice mentioned above.

In the phase-1 training process, gradient calculation in the back-propagation algorithm  \cite{lecun1989backpropagation} is the aggregate of the gradient calculated for each activation function. This gives a regularized effect on the gradient and provides equal opportunity for all weight parameters to optimize. Phase-2 training is the same as the standard training of the model with only one non-linear activation function (ReLU). Please note that a mixture of activation functions is used in only Phase-1 training, while Phase-2 has only ReLU at every layer. Further details are provided in the experimental section.

\section{Experiments and Results }
In this section, we have evaluated the performance of the proposed approach on classification and detection task. Our experimentation uses the state-of-art CNN architectures such as ResNet \cite{he2016deep}, and VGG-16 \cite{simonyan2014very} for various activation functions. All these models are trained over three standard benchmark datasets: CIFAR10, CIFAR100 \cite{krizhevsky2009learning} and ImageNet \cite{krizhevsky2012imagenet} dataset. We have also performed experiments using SSD \cite{liu2016ssd} on PASCAL VOC \cite{everingham2015pascal} for object detection. 

In these experiments, we have used four most prominently used non-linear activation functions (ReLU, PReLU, ELU, and SoftPlus). We have preferred PReLU to resolve the issue of dying ReLU; however, someone may also prefer Leaky ReLU. ELU has an exponential function for negative input, which is contrary to ReLU. This behavior of ELU pushes the mean to the neighborhood of zero, similar to the case of batch normalization \cite{ioffe2015batch}. This shift of mean toward the vicinity of zero accelerate the training of network (fast convergence). ELU also guarantees more robustness towards the noise. These are the few reasons we have selected ELU in the mixture of activation functions. The behavior of ReLU and Softplus \cite{zheng2015improving} is almost similar,  excluding near the periphery of zero, where the softplus is differentiable and smooth. Softplus has privilege over ReLU due to differentiability in the entire domain, and it saturates less.

The scope of activation functions depends on the problems. We have selected ReLU, PReLU, ELU, and SoftPlus, which are the most widely used non-linear activation functions in image classification and object detection problems.

In our experiments, baselines (using activation function ReLU/PReLU/ELU/SoftPlus) have been reproduced using a standard training procedure in PyTorch \cite{paszke2017automatic} framework. We trained these models using 300 and 90 epochs for CIFAR and ImageNet dataset respectively.

\subsection{CIFAR 10 and CIFAR 100}
The CIFAR-10 and CIFAR-100 \cite{krizhevsky2009learning} are the datasets having tiny natural images. CIFAR10 datasets have 10 different image classes, while CIFAR-100 datasets have 100 classes. There are 50,000 training images and 10,000 test images, where all images are RGB images with a dimension of 32x32.  

In the experiments with the CIFAR dataset, we perform standard data augmentation methods of random cropping to a size of $32\times32$ and random horizontal flipping. The optimization is performed using Stochastic Gradient Descent (SGD) algorithm with momentum $0.9$ and a minibatch size of $64$. In Phase-2 training, the initial learning rate is set to $0.1$, which is decreased by a factor of $5$ after every $50$ epoch. The models are trained from scratch for around $300$ epochs. For Phase-1 (Cooperative) training, we use only 20\% of epochs used in Phase-2 training with PReLU, ReLU, ELU, and SoftPlus activation functions (having equal weight). The learning rate is set to $0.1$ and is decreased by a factor of $5$ after every $10$ epoch in Phase-1 (Cooperative) training. For evaluation, the validation images are used. The results on the CIFAR-10/100 datasets for all the architectures have been reproduced in the PyTorch framework.

The results are shown in Table \ref{cifar10}, \ref{cifar100}. We observe a consistent improvement in accuracy for VGG-16 and ResNet-56 over CIFAR. The model trained with our proposed two-phase training procedure not only outperforms ReLU significantly but also other non-linear activation functions such as PReLU, ELU, and SoftPlus, as shown in Table \ref{cifar10}, \ref{cifar100}.

\begin{table}[t]
\begin{center}
\scalebox{.9}{
\begin{tabular}{lcc}
\toprule
Model & Activation Function & Accuracy(\%) \\
\midrule
VGG-16 (Baseline)& ReLU & 93.6 \\
VGG-16 (Baseline)& PReLU & 93.7\\
VGG-16 (Baseline)& SoftPlus & 90.5 \\
VGG-16 (Baseline)& ELU & 92.3 \\
\textbf{VGG-16 (Ours)} & \textbf{Mix (ReLU)} & \textbf{94.2} \\
\midrule
ResNet-56 (Baseline)& ReLU & 93.5 \\
ResNet-56 (Baseline)& PReLU & 94.0 \\
ResNet-56 (Baseline)& SoftPlus & 92.0 \\
ResNet-56 (Baseline)& ELU & 92.0 \\
\textbf{ResNet-56 (Ours)} & \textbf{Mix (ReLU)} & \textbf{ 94.4} \\
\bottomrule
\end{tabular}}
\vspace{2pt}
\caption{Classification accuracies for CIFAR 10. All baselines have been reproduced using corresponding activation function in PyTorch framework. Activation Function: Mix (ReLU) means model is trained using a mixture of Activation Functions in Phase-1 (Cooperative) training and then in Phase-2 model is trained using ReLU Activation Function.}
\vspace{-10pt}
\label{cifar10}
\end{center}
\end{table}

\begin{table}[t]
\begin{center}
\scalebox{.9}{
\begin{tabular}{lcc}
\toprule
Model & Activation Function & Accuracy(\%) \\
\midrule
VGG-16 (Baseline)& ReLU & 72.0 \\
VGG-16 (Baseline)& PReLU & 72.5 \\
VGG-16 (Baseline)& SoftPlus & 64.2 \\
VGG-16 (Baseline)& ELU & 66.9  \\
\textbf{VGG-16 (Ours)} & \textbf{Mix (ReLU)} & \textbf{74.0} \\
\midrule
ResNet-56 (Baseline)& ReLU & 71.6 \\
ResNet-56 (Baseline)& PReLU & 71.9 \\
ResNet-56 (Baseline)& SoftPlus & 69.3 \\
ResNet-56 (Baseline)& ELU & 69.5  \\
\textbf{ResNet-56 (Ours)} & \textbf{Mix (ReLU)} & \textbf{73.1} \\
\bottomrule
\end{tabular}}
\vspace{2pt}
\caption{Classification accuracies for CIFAR 100. All baselines have been reproduced using corresponding activation function in PyTorch framework. Activation Function: Mix (ReLU) means model is trained using a mixture of Activation Functions in Phase-1 (Cooperative) training and then in Phase-2 model is trained using ReLU Activation Function.}
\vspace{-10pt}
\label{cifar100} 
\end{center}
\end{table} 

\begin{table}[t]
\begin{center}
\scalebox{.9}{
\begin{tabular}{lcc}
\toprule
Model & Activation Function & Accuracy(\%) \\
\midrule
AlexNet (Baseline) & ReLU & 56.6 \\
AlexNet (Baseline) & PReLU & 56.9 \\
AlexNet (Baseline) & SoftPlus & 55.2 \\
AlexNet (Baseline) & ELU & 56.6 \\
\textbf{AlexNet (Ours)} & \textbf{Mix (ReLU)} & \textbf{57.2} \\
\midrule
ResNet-18 (Baseline) & ReLU & 69.8 \\
ResNet-18 (Baseline) & PReLU & 69.1 \\
ResNet-18 (Baseline) & SoftPlus & 68.8 \\
ResNet-18 (Baseline) & ELU & 68.2  \\
\textbf{ResNet-18 (Ours)} & \textbf{Mix (ReLU)} & \textbf{70.8} \\

\bottomrule
\end{tabular}}
\vspace{2pt}
\caption{Classification accuracies for ImageNet. The  accuracy is  reported  over validation set using  1-crop  setting (https://pytorch.org/docs/stable/torchvision/models.html).}
\vspace{-25pt}
\label{imagenet} 
\end{center}
\end{table}

\begin{table}[t]
\begin{center}
\scalebox{.9}{
\begin{tabular}{lcc}
\toprule
\textbf{Class} & \textbf{SSD (Baseline) AP} &\textbf{SSD Mix (ReLU) AP} \\
\midrule
\textbf{aero} & 80.40 &  82.53\\
\textbf{bike} & 82.95 & 82.54\\ 
\textbf{bird} &  74.62 & 77.02\\
\textbf{boat} & 71.61 & 72.45\\
\textbf{bottle} & 50.49 & 51.36\\
\textbf{bus} & 86.04 & 85.57\\
\textbf{car} & 86.55 & 86.28\\
\textbf{cat} & 88.02 & 86.91\\
\textbf{chair} & 60.88 & 63.23\\
\textbf{cow} & 83.10 & 81.58\\
\textbf{table} & 77.87 & 78.35\\
\textbf{dog} & 85.55 & 84.06\\
\textbf{horse} & 86.68 & 87.79\\
\textbf{mbike} & 84.14 & 85.85\\
\textbf{person} &  78.26 & 79.29 \\
\textbf{plant} & 50.44 & 52.82\\
\textbf{sheep} & 74.28 & 78.13\\
\textbf{sofa} &80.03 & 80.72\\ 
\textbf{train} & 85.88 & 87.28\\ 
\textbf{tv} & 75.49 & 77.15\\
\midrule
\textbf{mAP} & 77.16 &  \textbf{78.05} \\
\bottomrule
\end{tabular}}
\vspace{2pt}
\caption{AP for each class with Baseline SSD-300 (using standard training schedule) and SSD-300 Mix (ReLU) using our proposed training schedule on VOC2007 test dataset. Training data, 07+12 is the union of the VOC2007 and VOC2012 trainval dataset.}
\vspace{-25pt}
\label{tabssdvoc}
\end{center}
\end{table}

\subsection{ImageNet}
ImageNet dataset \cite{krizhevsky2012imagenet} contains 1000 classes, each category roughly having 1000 images. The dataset contains about 1.2 million training images, 50,000 validation images, and 100,000 test images (with no labels). The training is performed on training data, whereas all evaluations are performed on the validation set. 

For ImageNet experiments, we perform standard data augmentation methods of random cropping to a size of $224\times224$ and random horizontal flipping. For optimization, Stochastic Gradient Descent (SGD) is used with momentum $0.9$ and a minibatch size of $256$. For Phase-2 training, the initial learning rate is set to $0.1$, which is decreased by a factor of $10$ after every $30$ epoch. The models are trained for around $90$ epochs. The evaluation is done over validation images are subjected to center cropping of size $224\times224$. For Phase-1 (Cooperative) training, we use 20\% of epochs used in Phase-2 training with PReLU, ReLU, ELU, and SoftPlus activation functions (having equal weight). The learning rate is set to $0.1$ and is decreased by a factor of $10$ after every $5$ epoch in Phase-1 (Cooperative) training.

The results are shown in Table \ref{imagenet}. We have consistent improvement in accuracy for AlexNet \cite{krizhevsky2012imagenet}, ResNet-18 \cite{he2016deep} over ImageNet dataset. The model trained with our proposed two-phase training procedure not only significantly outperforms ReLU but also other non-linear activation function such as PReLU, ELU, and SoftPlus (Table \ref{imagenet}).

\subsection{PASCAL VOC}

We have performed experiments on the SSD model over  PASCAL VOC \cite{everingham2015pascal} dataset to validate our proposed approach for the object detection task. In this experiment, we follow the same experimental setting and training schedule, as described in \cite{liu2016ssd} for Phase-2 training. For Phase-1 (Cooperative) training, we have used 20\% iterations of Phase-2 training with PReLU, ReLU, ELU, and SoftPlus activation functions (having equal weight). The  SSD \cite{liu2016ssd} detection model is a feed-forward convolutional network that produces a collection of fixed-size bounding boxes and predicts classification scores to represent the presence of object class instances in these boxes, followed by a non-maximum suppression which produces the final detections.

As shown in Table \ref{tabssdvoc}, our proposed approach is not limited to classification but also works well on object detection task. We have a significant improvement  (approx. 1\%) in mAP as compare to baseline by using our training procedure.

\begin{table}[t]
\begin{center}
\scalebox{.82}{
\begin{tabular}{lcc}
\toprule
Model & Activation Function & Accuracy(\%) \\
\midrule
ResNet-56 (Baseline)& ReLU & 93.5 \\
ResNet-56 (Baseline)& PReLU & 94.0 \\
ResNet-56 (Baseline)& SoftPlus & 92.0 \\
ResNet-56 (Baseline)& ELU & 92.0 \\
\midrule
ResNet-56 (Baseline-TPT)& ReLU & 93.6 \\
ResNet-56 (Baseline-TPT)& PReLU & 94.0 \\
ResNet-56 (Baseline-TPT)& SoftPlus & 92.0 \\
ResNet-56 (Baseline-TPT)& ELU & 92.1 \\
\midrule
ResNet-56 & WNLA & 40.0 \\
\midrule
ResNet-56 & Mixture &  94.3\\
ResNet-56 & Mix (PReLU) & 94.1 \\
ResNet-56 & Mix (SoftPlus) &  93.1\\
ResNet-56 & Mix (ELU) &  93.2\\
\midrule
\textbf{ResNet-56 (Ours)} & \textbf{Mix (ReLU)} & \textbf{ 94.4} \\
\bottomrule
\end{tabular}}
\vspace{2pt}
\caption{ Classification accuracies for ResNet-56 on CIFAR 10. Activation Function: Mixture means model is trained completely using a mixture (ReLU, PReLU, ELU, and SoftPlus) of Activation Functions. Activation Function: Mix ($\gamma$) means model is trained using a mixture of Activation Functions in Phase-1 (Cooperative) training and then in Phase-2 model is trained using $\gamma$ Activation Function. Baseline-TPT means baseline uses Two Phase Training.}
\vspace{-12pt}
\label{ablation}
\end{center}
\end{table}

\begin{table}[t]
\begin{center}
\scalebox{.85}{
\begin{tabular}{lcc}
\toprule
\#Epochs in Phase-1  & Activation Function & Accuracy(\%) \\
\midrule
10\% & Mix (ReLU) &  94.03 \\
\textbf{20\%} & Mix (ReLU) &  \textbf{94.40} \\
30\% & Mix (ReLU) &  94.40 \\
40\% & Mix (ReLU) &  94.42 \\
\bottomrule
\end{tabular}}
\vspace{2pt}
\caption{The table shows results for changing the number of Epochs in Phase-1 training for ResNet-56 model on CIFAR 10.}
\vspace{-25pt}
\label{ablationepoch}
\end{center}
\end{table}

\section{Ablation Studies}

As shown in Table \ref{ablation}, if we train a model without any non-linear activation function (WNLA), the performance of the model degrades massively since ResNet-56 is a deep CNN model which is very hard to optimize without any non-linear activation function. There is a performance boost in the case of Mix (SoftPlus) and Mix (ELU) from their respective baselines, but the overall performance scores are significantly lower than Mix (ReLU). The key reason for the significant difference in the performance can be due to the sparse behavior of (ReLU) activation function in Phase-2 training. This sparsity is often desirable in the deep network due to better predictive power and less overfitting/noise.

Although Mixture and Mix (ReLU) have similar performance, Mix (ReLU) should be preferred because of the following reasons:

\begin{itemize}
    \item Mixture will occupy more feature maps memory as compare to Mix (ReLU) at run time because separate feature maps will be generated for every non-linear activation function. Mix (ReLU) will not increase in feature maps memory at a run time (GPU Memory).
    \item Mixture will add some delay at the inference time as compare to Mix (ReLU) due to extra calculations performed by multiple activation functions.
\end{itemize}

Hence, Mix (ReLU) is the most suitable combination while considering all the other possibilities. Our proposed approach uses Two-Phase Training (TPT), where Phase-1 training uses 20\% of Phase-2 epochs. Therefore, one may think that performance improvement is due to more training epochs (20\%). Hence we also train baselines with the same Two-Phase Training schedule (Baseline-TPT), except in Phase-1, only a single activation function is used. As shown in Table \ref{ablation}, Baseline-TPT has a similar performance as the baseline. From Table~\ref{ablation}, we can conclude that performance improvement is not due to more training epochs (20\%) in Two-Phase Training (TPT) but because of \textit{cooperative initialization} in Phase-1.

We also conduct an ablation to decide the number of Epochs in Phase-1 training. For Phase-1 (Cooperative) training, we use 10-40\% of epochs used in standard (Phase-2) training. As shown in Table~\ref{ablationepoch}, 20\% of Phase-2 epochs in Phase-1 (Cooperative) training is the most suitable choice because it gives significant performance improvement with only 20\% increase in overall training time.

\section{Visualizing Last Layer Features on MNIST}

\begin{figure*}[t]
    \centering
    \includegraphics[scale=0.27]{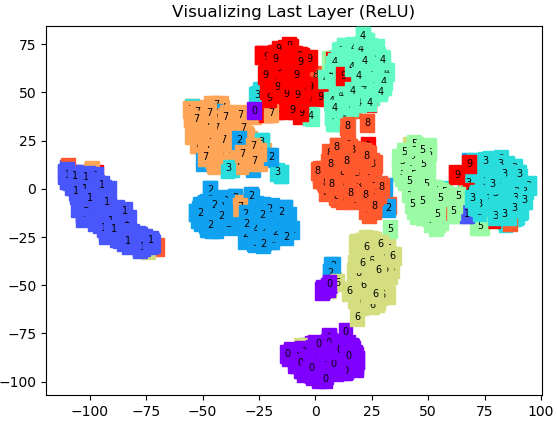}
    \includegraphics[scale=0.27]{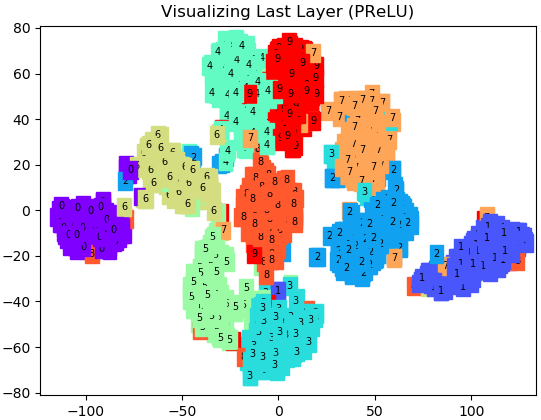}
    \includegraphics[scale=0.27]{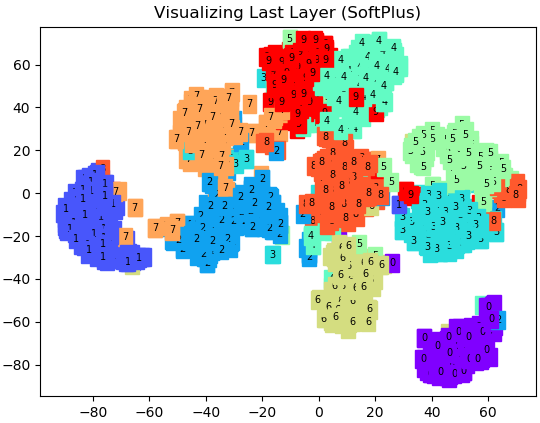}
    \includegraphics[scale=0.27]{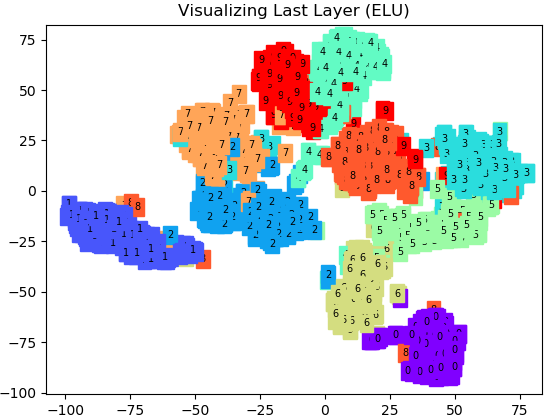}
    \includegraphics[scale=0.27]{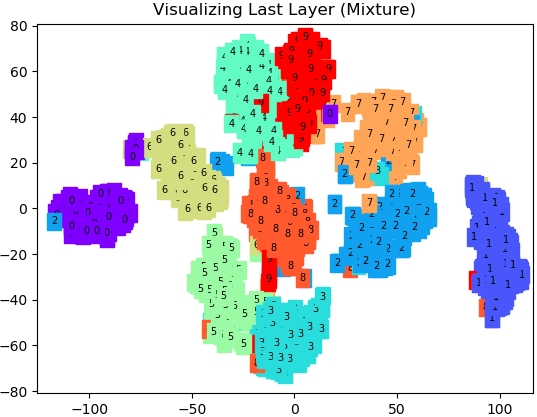}
    \includegraphics[scale=0.37]{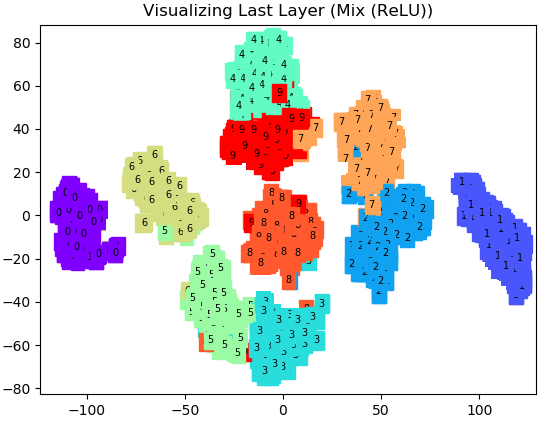}
\caption{Two-dimensional t-SNE visualization of trained flatten layer features for LeNet-like network on MNIST (Best viewed in color).}
\label{tsne}
\end{figure*}

\begin{figure*}[t]
    \centering
    \includegraphics[scale=0.28]{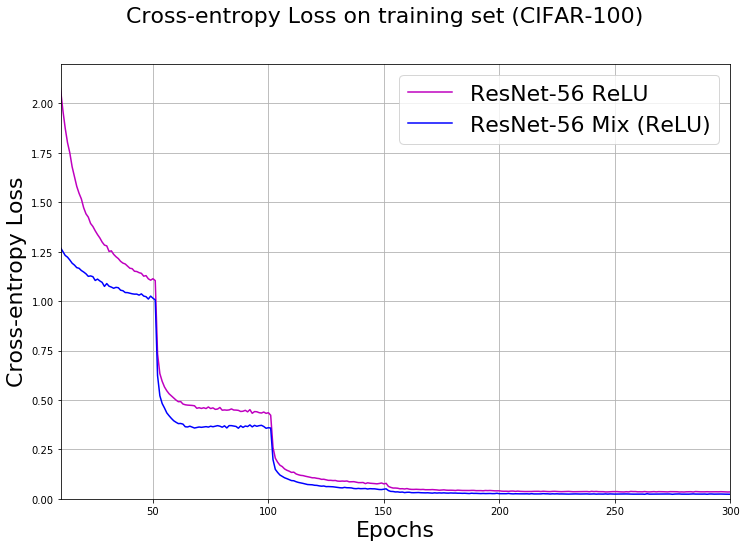}
    \includegraphics[scale=0.28]{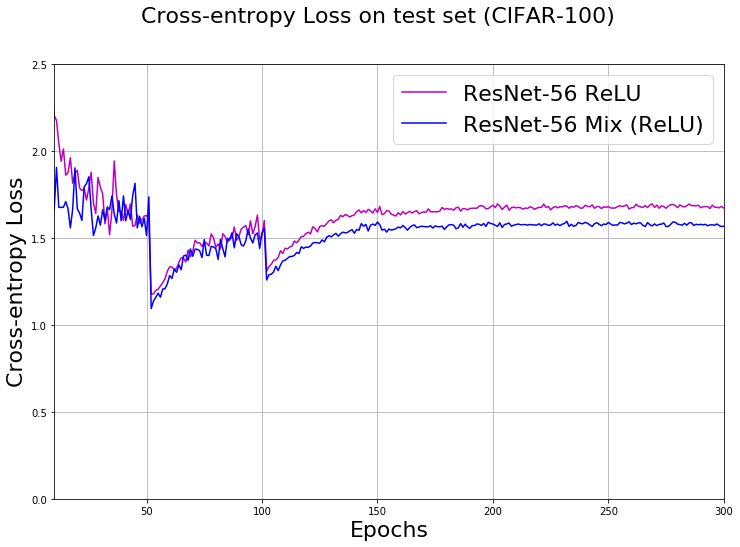}
\caption{ResNet-56 ReLU and ResNet-56 Mix (ReLU) training and test losses over  CIFAR-100 dataset. ResNet-56 Mix (ReLU)  exhibits improved optimisation/convergence characteristics and produces significant gain in performance (Best viewed in color).}
\vspace{-12pt}
\label{fig:convergence}
\end{figure*}

In the Figure \ref{tsne}, we are plotting t-SNE \cite{maaten2008visualizing} plots for LeNet-like network on MNIST \cite{lecun-mnisthandwrittendigit-2010} dataset to visualize the features learned for various non linear activation functions. The LeNet-like network contains two convolutional layers and one fully-connected layer. The convolutional layers have 5x5 kernel size while the first and second convolutional layer consists of twenty and thirty number of filters respectively.

The t-Distributed Stochastic Neighbor Embedding (t-SNE) is a dimensionality reduction technique that is heavily used to visualize the high-level features learned by CNN. The t-SNE is a commonly used technique to visualize feature representations in high-dimensional data into a space of two or three dimensions. From Figure \ref{tsne}, we can visualize the two-dimensional embeddings of the last layer. The features corresponding to Mix (Relu) are more separable than remaining other embeddings. The features corresponding to Mix (ReLU) are well separated and discriminated enough, as shown using corresponding representative points in Figure \ref{tsne}. 

\section{Analysis}
This section focused on the analysis of performance gain from two different perspectives. The first perspective focus on investigating the convergence of Mix (ReLU)  and ReLU on the ResNet-56 architecture. The convergence speed of Mix (ReLU) is much faster as compared to ReLU, which can be inferred in Figure-\ref{fig:convergence}. The second perspective for investigation is based on over-fitting aspects of the models where Mix (ReLU) is more robust compared to ReLU based on the empirical results.

The investigation is performed on the CIFAR-100 dataset using the ResNet-56 model. We used standard data augmentation techniques such as random horizontal flipping and random cropping to size $32\times32$. The optimization is performed using Stochastic Gradient Descent (SGD) with momentum set to $0.9$ and weight decay as 0.0001. The minibatch size of $64$ is selected to perform experiments. The initial learning rate is taken as $0.1$, which is then decreased by a factor of $5$ after every $50$ epoch. The models are trained from scratch for around $300$ epochs.

\subsection{Effect of using Mix (ReLU) on convergence}

\begin{figure*}[t]
    \centering
    \includegraphics[scale=0.3]{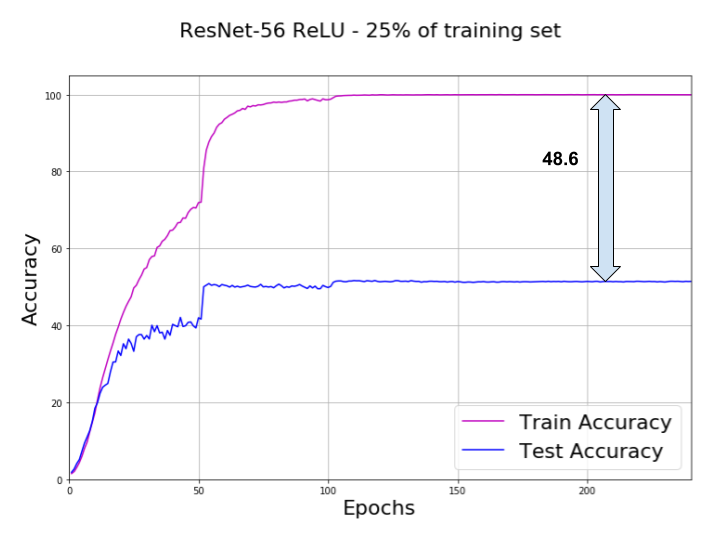}
    \includegraphics[scale=0.3]{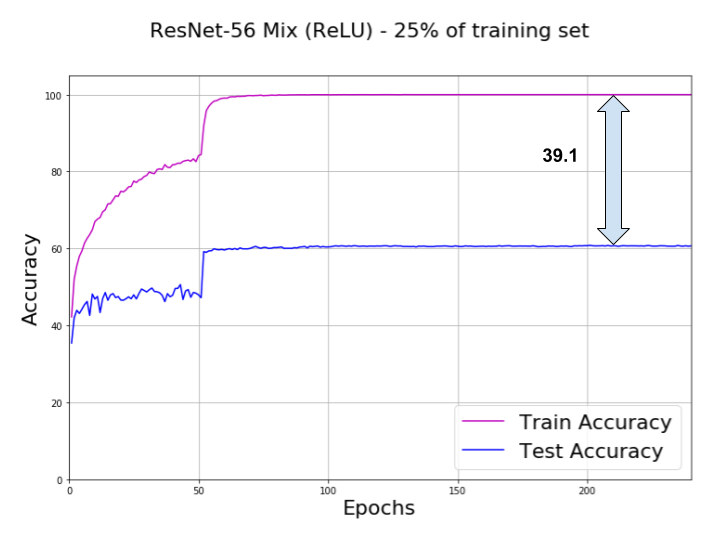}
\caption{Training  curves  for  ResNet-56 ReLU and ResNet-56 Mix (ReLU) models  over  CIFAR-100 dataset (Best viewed in color).}

\label{fig:overfitting}
\end{figure*}

\begin{table*}[t]
    \centering
        \addtolength{\tabcolsep}{2pt}
        \begin{tabular}{|l| c| c|c|} 
            \hline
            Method & Activation Function & Train Accuracy(\%)  & Test Accuracy(\%) \\
            \hline\hline
            ResNet-56 Baseline (100\% of training set) & ReLU & 99.4  & 71.6 \\
            ResNet-56 (100\% of training set) & \textbf{Mix (ReLU)} & 99.6  & \textbf{73.1} \\
            \hline\hline
            ResNet-56 Baseline (25\% of training set) & ReLU &  99.9 &   51.3 \\
            ResNet-56 (25\% of training set) & \textbf{Mix (ReLU)} &  99.9 &  \textbf{60.8} \\
            \hline
        \end{tabular}
    \caption{The table shows the results for ResNet-56 ReLU and ResNet-56 Mix (ReLU) over CIFAR-100 dataset in different setups.}
    \vspace{-5pt}
    \label{tab:overfitting}
\end{table*}

We analyzed the convergence rate of Mix (ReLU) based model, and we found that the convergence using Mix (ReLU) is slightly better compared to ReLU, which can be inferred from their respective curves in Figure \ref{fig:convergence}. The two graphs of cross-entropy losses vs. the number of epochs for training and test set in  Figure-\ref{fig:convergence}, shows that the dropping rate of cross-entropy losses is quite higher as compared to the loss corresponding to ReLU on the training set in the experimental results shown in Figure-\ref{fig:convergence}.

\subsection{Effect of Mix (ReLU) on over fitting}
Our method utilize multiple activation functions in the initial few training epochs of the deep network.  The gradient from these multiple activation functions gets accumulated and gave a \textit{regularization effect} to the gradient flow in the model corresponding to the complete range of input data (negative and positive ). This aggregation of gradients also helps in the regularization of weight parameters while updating, which in effect knock-down the chances of over-fitting issues. In support of our hypothesis, we have performed some experiments in two different scenarios. In the first scenario, experiments are performed out over the complete dataset (100 \% of training data), while the second scenario of experiments is performed on only 25 \% of the training data.  

The first scenario of experiments is performed on the CIFAR-100 dataset using ResNet-56 architecture. The first experiment with only ReLU achieved an accuracy of 71.6\% as presented in Table-\ref{tab:overfitting} while the second experiment uses Mix (ReLU) and achieve 73.1\% accuracy. From Table-\ref{tab:overfitting}, we can infer that Mix (ReLU) is more robust to overfitting as the difference between test and training accuracy is 26.5, which is lesser than the difference between test and training accuracy of ReLU (27.8). In these experiments, the difference between test and training accuracy is not that much significant for ReLU and Mix (ReLU), hence in the second scenario, we have chosen only 25\% samples from the training images of CIFAR-100 dataset to perform various experiments.

The second scenario of experiments is performed to train the ResNet-56 ReLU model using only 25\% train samples, and we achieve accuracy of 99.9\% and 51.3\% corresponding to train and test data respectively. The ResNet-56 Mix (ReLU), on the other hand, achieves 99.9\% and 60.8\%  of training and test accuracy respectively. 

From Figure-\ref{fig:overfitting}, We can conclude that the Mix (ReLU) is less prone to overfitting issues, as the difference between test and training accuracy is 39.1 while this difference is quite higher than that of ReLU, i.e., 48.6. 

\section{Conclusion}
We propose a Cooperative Initialization for training deep networks to improve the performance. We have shown experimentally that a mixture of the non-linear activation function is beneficial for CNN in the initial phase of training, where we start from random initialization. Initially, multiple activation functions regularize the gradient flow corresponding to the positive and negative input of activation functions, thereby improving the update of weight parameters, which is very crucial at the initial stage. Our experimental results show that the proposed approach improves the performances of state-of-the-art networks. Our proposed approach also helps in reducing the overfitting problem and does not increase the number of parameters, inference (test) time in the final model while improving the performance. Therefore, cooperative initialization is a promising approach to improve the feature representation and performance of deep networks.

{\small
\bibliographystyle{ieee}
\bibliography{egbib}

\begin{thebibliography}{10}\itemsep=-1pt

\bibitem{agostinelli2014learning}
F.~Agostinelli, M.~Hoffman, P.~Sadowski, and P.~Baldi.
\newblock Learning activation functions to improve deep neural networks.
\newblock {\em arXiv preprint arXiv:1412.6830}, 2014.

\bibitem{bishop1995neural}
C.~M. Bishop et~al.
\newblock {\em Neural networks for pattern recognition}.
\newblock Oxford university press, 1995.

\bibitem{clevert2015fast}
D.-A. Clevert, T.~Unterthiner, and S.~Hochreiter.
\newblock Fast and accurate deep network learning by exponential linear units
  (elus).
\newblock {\em arXiv preprint arXiv:1511.07289}, 2015.

\bibitem{collobert2008unified}
R.~Collobert and J.~Weston.
\newblock A unified architecture for natural language processing: Deep neural
  networks with multitask learning.
\newblock In {\em Proceedings of the 25th international conference on Machine
  learning}, pages 160--167. ACM, 2008.

\bibitem{everingham2015pascal}
M.~Everingham, S.~A. Eslami, L.~Van~Gool, C.~K. Williams, J.~Winn, and
  A.~Zisserman.
\newblock The pascal visual object classes challenge: A retrospective.
\newblock {\em International journal of computer vision}, 111(1):98--136, 2015.

\bibitem{goodfellow2013maxout}
I.~J. Goodfellow, D.~Warde-Farley, M.~Mirza, A.~Courville, and Y.~Bengio.
\newblock Maxout networks.
\newblock {\em arXiv preprint arXiv:1302.4389}, 2013.

\bibitem{graves2013speech}
A.~Graves, A.-r. Mohamed, and G.~Hinton.
\newblock Speech recognition with deep recurrent neural networks.
\newblock In {\em ICASSP}, 2013.

\bibitem{he2015delving}
K.~He, X.~Zhang, S.~Ren, and J.~Sun.
\newblock Delving deep into rectifiers: Surpassing human-level performance on
  imagenet classification.
\newblock In {\em Proceedings of the IEEE international conference on computer
  vision}, pages 1026--1034, 2015.

\bibitem{he2016deep}
K.~He, X.~Zhang, S.~Ren, and J.~Sun.
\newblock Deep residual learning for image recognition.
\newblock In {\em Proceedings of the IEEE conference on computer vision and
  pattern recognition}, pages 770--778, 2016.

\bibitem{hinton2009replicated}
G.~E. Hinton and R.~R. Salakhutdinov.
\newblock Replicated softmax: an undirected topic model.
\newblock In {\em Advances in neural information processing systems}, pages
  1607--1614, 2009.

\bibitem{ioffe2015batch}
S.~Ioffe and C.~Szegedy.
\newblock Batch normalization: Accelerating deep network training by reducing
  internal covariate shift.
\newblock {\em arXiv preprint arXiv:1502.03167}, 2015.

\bibitem{krizhevsky2009learning}
A.~Krizhevsky.
\newblock Learning multiple layers of features from tiny images.
\newblock Technical report, Citeseer, 2009.

\bibitem{krizhevsky2012imagenet}
A.~Krizhevsky, I.~Sutskever, and G.~E. Hinton.
\newblock Imagenet classification with deep convolutional neural networks.
\newblock In {\em Advances in neural information processing systems}, pages
  1097--1105, 2012.

\bibitem{lecun2015deep}
Y.~LeCun, Y.~Bengio, and G.~Hinton.
\newblock Deep learning.
\newblock {\em nature}, 2015.

\bibitem{lecun1989backpropagation}
Y.~LeCun, B.~Boser, J.~S. Denker, D.~Henderson, R.~E. Howard, W.~Hubbard, and
  L.~D. Jackel.
\newblock Backpropagation applied to handwritten zip code recognition.
\newblock {\em Neural computation}, 1(4):541--551, 1989.

\bibitem{lecun-mnisthandwrittendigit-2010}
Y.~LeCun and C.~Cortes.
\newblock {MNIST} handwritten digit database.
\newblock 2010.

\bibitem{liu2016ssd}
W.~Liu, D.~Anguelov, D.~Erhan, C.~Szegedy, S.~Reed, C.-Y. Fu, and A.~C. Berg.
\newblock Ssd: Single shot multibox detector.
\newblock In {\em European conference on computer vision}, pages 21--37.
  Springer, 2016.

\bibitem{maas2013rectifier}
A.~L. Maas, A.~Y. Hannun, and A.~Y. Ng.
\newblock Rectifier nonlinearities improve neural network acoustic models.

\bibitem{maaten2008visualizing}
L.~v.~d. Maaten and G.~Hinton.
\newblock Visualizing data using t-sne.
\newblock {\em Journal of machine learning research}, 2008.

\bibitem{mazumder2019cpwc}
P.~Mazumder, P.~Singh, and V.~Namboodiri.
\newblock Cpwc: Contextual point wise convolution for object recognition.
\newblock {\em arXiv preprint arXiv:1910.09643}, 2019.

\bibitem{mcculloch1943logical}
W.~S. McCulloch and W.~Pitts.
\newblock A logical calculus of the ideas immanent in nervous activity.
\newblock {\em The bulletin of mathematical biophysics}, 1943.

\bibitem{nair2010rectified}
V.~Nair and G.~E. Hinton.
\newblock Rectified linear units improve restricted boltzmann machines.
\newblock In {\em Proceedings of the 27th international conference on machine
  learning (ICML-10)}, pages 807--814, 2010.

\bibitem{paszke2017automatic}
A.~Paszke, S.~Gross, S.~Chintala, G.~Chanan, E.~Yang, Z.~DeVito, Z.~Lin,
  A.~Desmaison, L.~Antiga, and A.~Lerer.
\newblock Automatic differentiation in pytorch.
\newblock 2017.

\bibitem{simonyan2014very}
K.~Simonyan and A.~Zisserman.
\newblock Very deep convolutional networks for large-scale image recognition.
\newblock {\em arXiv preprint arXiv:1409.1556}, 2014.

\bibitem{singh2019falf}
P.~Singh, V.~S.~R. Kadi, and V.~P. Namboodiri.
\newblock Falf convnets: Fatuous auxiliary loss based filter-pruning for
  efficient deep cnns.
\newblock {\em Image and Vision Computing}, page 103857, 2019.

\bibitem{singh2019stability}
P.~Singh, V.~S.~R. Kadi, N.~Verma, and V.~P. Namboodiri.
\newblock Stability based filter pruning for accelerating deep cnns.
\newblock In {\em 2019 IEEE Winter Conference on Applications of Computer
  Vision (WACV)}, pages 1166--1174. IEEE, 2019.

\bibitem{singh2019multi}
P.~Singh, R.~Manikandan, N.~Matiyali, and V.~Namboodiri.
\newblock Multi-layer pruning framework for compressing single shot multibox
  detector.
\newblock In {\em 2019 IEEE Winter Conference on Applications of Computer
  Vision (WACV)}, pages 1318--1327. IEEE, 2019.

\bibitem{singh2019accuracy}
P.~Singh, P.~Mazumder, and V.~P. Namboodiri.
\newblock Accuracy booster: Performance boosting using feature map
  re-calibration.
\newblock {\em arXiv preprint arXiv:1903.04407}, 2019.

\bibitem{singh2018leveraging}
P.~Singh, V.~K. Verma, P.~Rai, and V.~P. Namboodiri.
\newblock Leveraging filter correlations for deep model compression.
\newblock {\em arXiv preprint arXiv:1811.10559}, 2018.

\bibitem{singh2019hetconvijcv}
P.~Singh, V.~K. Verma, P.~Rai, and V.~P. Namboodiri.
\newblock Hetconv: Beyond homogeneous convolution kernels for deep cnns.
\newblock {\em International Journal of Computer Vision}, pages 1--21, 2019.

\bibitem{singh2019hetconv}
P.~Singh, V.~K. Verma, P.~Rai, and V.~P. Namboodiri.
\newblock Hetconv: Heterogeneous kernel-based convolutions for deep cnns.
\newblock In {\em Proceedings of the IEEE Conference on Computer Vision and
  Pattern Recognition}, pages 4835--4844, 2019.

\bibitem{singh2019play}
P.~Singh, V.~K. Verma, P.~Rai, and V.~P. Namboodiri.
\newblock Play and prune: Adaptive filter pruning for deep model compression.
\newblock {\em International Joint Conference on Artificial Intelligence
  (IJCAI)}, 2019.

\bibitem{szegedy2015going}
C.~Szegedy, W.~Liu, Y.~Jia, P.~Sermanet, S.~Reed, D.~Anguelov, D.~Erhan,
  V.~Vanhoucke, and A.~Rabinovich.
\newblock Going deeper with convolutions.
\newblock In {\em Proceedings of the IEEE conference on computer vision and
  pattern recognition}, pages 1--9, 2015.

\bibitem{trottier2017parametric}
L.~Trottier, P.~Gigu, B.~Chaib-draa, et~al.
\newblock Parametric exponential linear unit for deep convolutional neural
  networks.
\newblock In {\em ICMLA}, 2017.

\bibitem{xu2015empirical}
B.~Xu, N.~Wang, T.~Chen, and M.~Li.
\newblock Empirical evaluation of rectified activations in convolutional
  network.
\newblock {\em arXiv preprint arXiv:1505.00853}, 2015.

\bibitem{zheng2015improving}
H.~Zheng, Z.~Yang, W.~Liu, J.~Liang, and Y.~Li.
\newblock Improving deep neural networks using softplus units.
\newblock In {\em IJCNN}.

\end{thebibliography}
}

\end{document}